\documentclass{article}
\usepackage[preprint]{neurips_2026}
\usepackage[utf8]{inputenc}
\usepackage[T1]{fontenc}
\usepackage{hyperref}
\usepackage{url}
\usepackage{booktabs}
\usepackage{amsfonts}
\usepackage{amsmath}
\usepackage{nicefrac}
\usepackage{microtype}
\usepackage{xcolor}
\usepackage{graphicx}
\usepackage{longtable}
\usepackage{array}
\usepackage{calc}
\usepackage{caption}
\usepackage{float}
\usepackage{textcomp}
\usepackage{upquote}
\usepackage{enumitem}
\usepackage{etoolbox}
\setlist{nosep,leftmargin=*}
\providecommand{\tightlist}{\setlength{\itemsep}{0pt}\setlength{\parskip}{0pt}}
\AtBeginEnvironment{longtable}{\scriptsize}
\title{The Convergence Gap: Instruction-Tuned Language Models Stabilize Later in the Forward Pass}
\author{Yifan Zhou\\
\normalfont\small University of California, Los Angeles\\
\normalfont\small \texttt{yifanz1207@gmail.com}\\
\normalfont\small Code: \href{https://github.com/yifan1207/convergence-gap-instruction-tuning}{\texttt{github.com/yifan1207/convergence-gap-instruction-tuning}}\\
\normalfont\small Artifact release: \href{https://github.com/yifan1207/convergence-gap-instruction-tuning/releases/tag/paper-artifacts-v1}{paper-artifacts-v1}\\
\normalfont\small Raw mirror: \url{gs://pt-vs-it-results/papers/convergence_gap/}}

\begin{document}
\maketitle
\begin{abstract}
Final outputs hide when a checkpoint commits to its next-token prediction. We introduce the convergence gap, a model-diffing diagnostic that decodes each layer's next-token distribution and measures its distance to the model's own final distribution. Across six paired pretrained and instruction-tuned checkpoints in native prompting regimes, instruction-tuned checkpoints remain farther from their final predictions later into the stack. The effect persists under endpoint-matched raw and tuned readouts, endpoint-free same-history checks, and fixed-history template replay. Matched-prefix interventions identify late MLP windows as the largest tested leverage point: late IT grafts into PT hosts increase late KL by +0.34 nats, while PT-late swaps into IT hosts reduce it by -0.51 nats; matched random late perturbations give only +0.003 versus +0.327 for the true late graft. A preselected Gemma case study provides behavior-facing plausibility for the same late swap, without serving as a benchmark claim. These results identify a robust prediction-dynamics signature of post-training: released instruction-following checkpoints tend to settle later, and late MLP computation is the strongest tested bidirectional handle on that delay under matched histories.
\end{abstract}

\section{Introduction}\label{introduction}

Model diffing needs diagnostics that compare paired checkpoints without reducing the comparison to final-output scores. A useful diagnostic should ask not only whether two checkpoints end at different predictions, but when each checkpoint settles on its own prediction during the forward pass. We introduce the convergence gap as such a diagnostic and use pretrained/instruction-tuned checkpoint pairs as the main test case.

We turn that question into a layerwise metric. For each layer, we decode a next-token distribution and compare it to the model's own final distribution. The resulting \textbf{convergence gap} asks how far the current layer is from the prediction that the model will finally output. A larger late gap means the model is still changing its mind later in the stack.

Across six released pretrained/instruction-tuned checkpoint pairs, instruction-tuned models show a larger late convergence gap. This is not just the familiar observation that late layers sharpen logits. We compare each model to its own endpoint, match endpoint confidence and entropy, and also use endpoint-free same-history checks that do not compare to the final distribution at all. The gap remains.

\begin{quote}\small
\textbf{Result in one place.}

\begin{itemize}
\tightlist
\item
  Instruction-tuned checkpoints show a larger late convergence gap across six PT/IT pairs.
\item
  Endpoint matching preserves the gap: raw +0.425 nats and tuned +0.762 nats.
\item
  Endpoint-free checks agree: adjacent JS +0.052 and future top-1 flips +0.203.
\item
  Matched-prefix MLP interventions identify late windows as the largest tested leverage point: late IT graft +0.34 nats and late PT swap -0.51 nats.
\item
  Matched random late perturbations do not reproduce the effect: +0.003 nats versus true +0.327.
\item
  A preselected Gemma behavior-facing case study shows native IT is preferred over the late PT swap on assistant-register prompts in 94.7\% of pairwise comparisons; this is a high-signal bridge, not a benchmark claim.
\end{itemize}
\end{quote}

The paper makes three contributions.

\begin{enumerate}
\def\labelenumi{\arabic{enumi}.}
\tightlist
\item
  \textbf{A prediction-dynamics target for post-training.} The convergence gap measures when a model settles toward its own final next-token prediction, making stabilization through depth a model-diffing object.
\item
  \textbf{Endpoint and history controls.} Raw/tuned endpoint-matched readouts and endpoint-free same-history statistics test whether the gap is only a confidence, probe, or endpoint artifact.
\item
  \textbf{Matched-prefix MLP leverage.} Symmetric graft/swap experiments and matched random late perturbations show that late MLP windows are the largest tested bidirectional handle on delayed stabilization.
\end{enumerate}

A preselected Gemma case study then asks whether the same late-window handle can matter in generation. It is deliberately secondary: it checks behavior-facing plausibility for the clearest family, not whether the convergence gap predicts benchmark performance or instruction following across tasks.

Scope matters. We do not claim a full mechanism for instruction following, a deployment-level behavior estimate across tasks, or a universal steering direction. The paper is about layerwise prediction dynamics under matched histories in released PT/IT checkpoint pairs, with one behavior-facing Gemma case study.

This paper deliberately does not test whether a late-stack effect ports across upstream states from another checkpoint at a selected token. Its primary object is the layerwise stabilization trajectory itself.

\section{Setup}\label{setup}

\subsection{Models and Histories}\label{models-and-histories}

We compare six paired pretrained (PT) and instruction-tuned (IT) checkpoints: Gemma 3 4B, Llama 3.1 8B, Qwen 3 4B, Mistral 7B, OLMo 2 7B, and DeepSeek-V2-Lite. The main discovery curves use all six pairs. Matched-prefix causal averages are reported on the dense families; DeepSeek is treated as a side case because MoE routing changes the interpretation of swapping dense MLP windows.

All comparisons are made under matched token histories. The discovery curves aggregate per-token layer readouts; the raw/tuned token-step counts range from 1.66e5 to 1.38e6 per checkpoint branch, with exact counts in Appendix A. For graft/swap interventions, the token prefix is held fixed while one MLP window is replaced. Bootstrap intervals are over prompt clusters where prompt-level rows are available; endpoint-matched controls additionally report matched-token retention and post-match balance.

\subsection{Convergence Gap}\label{convergence-gap}

Let \texttt{p\_l} be the next-token distribution decoded from layer \texttt{l}, and let \texttt{p\_L} be the model's own final next-token distribution. The layerwise convergence gap is:

\texttt{KL(p\_l\ \textbar{}\textbar{}\ p\_L)}.

This is a within-model quantity. It does not ask whether PT and IT have the same final distribution. It asks how far each checkpoint is from its own final prediction at each layer.

We use two companion checks. \textbf{Endpoint-matched convergence gap} matches PT and IT token steps by endpoint confidence, entropy, and top1/top2 margin before comparing late gaps. \textbf{Endpoint-free same-history checks} compare PT and IT native same-layer output distributions under identical teacher-forced token histories, avoiding each model's final endpoint.

\subsection{Matched-Prefix MLP Interventions}\label{matched-prefix-mlp-interventions}

Matched-prefix interventions use overlapping early, middle, and late MLP windows of roughly 40\% model depth, centered at the corresponding depth region. For example, Llama 3.1 8B uses layers 0-12, 9-21, and 19-31; exact per-family ranges are in Appendix B. On the PT host, \texttt{B\_early}, \texttt{B\_mid}, and \texttt{B\_late} replace a PT MLP window with the corresponding IT MLP window. On the IT host, \texttt{D\_early}, \texttt{D\_mid}, and \texttt{D\_late} replace an IT MLP window with PT. This gives a symmetric bidirectional leverage test for delayed stabilization under matched histories.

A separate random-control follow-up replaces the true late IT-minus-PT residual effect with matched random residual-projection perturbations. This tests whether any matched late perturbation recreates the convergence-gap signature.

\section{Results}\label{results}

\subsection{Instruction-Tuned Models Stay Farther From Their Final Predictions}\label{instruction-tuned-models-stay-farther-from-their-final-predictions}

The tuned-lens discovery view shows a broad late convergence gap across all six families. IT curves remain higher later in the stack, meaning the decoded distributions are farther from the final distribution at the same relative depth.

\begin{figure}[H]
\centering
\includegraphics[width=0.94\linewidth,height=0.22\textheight,keepaspectratio]{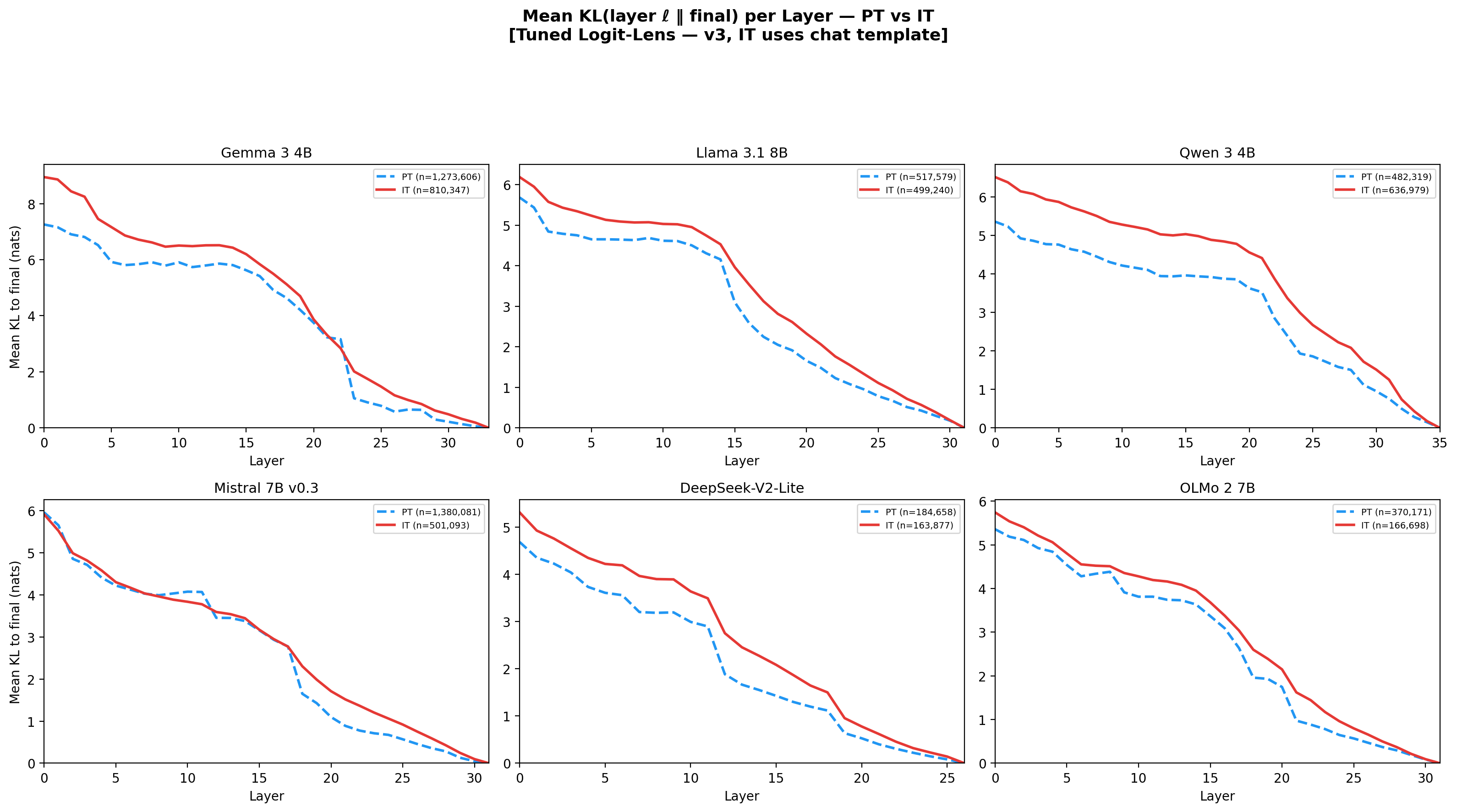}
\caption{Mean KL(layer to final) per layer for six PT/IT families. IT curves remain higher through much of the stack, indicating a broader convergence gap.}
\end{figure}

Endpoint matching is the strongest control for this observation. We match token steps on final confidence, entropy, and top1/top2 margin, then remeasure the late gap. The IT-minus-PT gap remains positive in both raw and tuned readouts.

{\def\LTcaptype{none} 
\begin{longtable}[]{@{}
  >{\raggedright\arraybackslash}p{(\linewidth - 4\tabcolsep) * \real{0.3000}}
  >{\raggedleft\arraybackslash}p{(\linewidth - 4\tabcolsep) * \real{0.4000}}
  >{\raggedright\arraybackslash}p{(\linewidth - 4\tabcolsep) * \real{0.3000}}@{}}
\toprule\noalign{}
\begin{minipage}[b]{\linewidth}\raggedright
Check
\end{minipage} & \begin{minipage}[b]{\linewidth}\raggedleft
Estimate
\end{minipage} & \begin{minipage}[b]{\linewidth}\raggedright
Interpretation
\end{minipage} \\
\midrule\noalign{}
\endhead
\bottomrule\noalign{}
\endlastfoot
Endpoint-matched raw late KL & \texttt{+0.425} nats \texttt{{[}+0.356,\ +0.493{]}} & raw-lens gap remains \\
Endpoint-matched tuned late KL & \texttt{+0.762} nats \texttt{{[}+0.709,\ +0.814{]}} & tuned-lens gap remains \\
Endpoint-free adjacent JS & \texttt{+0.052} JS \texttt{{[}+0.048,\ +0.057{]}} & IT has more remaining layer-to-layer movement \\
Endpoint-free future top-1 flips & \texttt{+0.203} flips \texttt{{[}+0.190,\ +0.215{]}} & IT changes top-1 later \\
\end{longtable}
}

The matching quality is clean: minimum matched retention is 0.796, maximum post-match standardized mean difference is 0.057, and malformed rate is zero. The convergence gap is therefore not explained away by endpoint confidence, entropy, or a single tuned-lens threshold.

\begin{figure}[H]
\centering
\includegraphics[width=0.94\linewidth,height=0.20\textheight,keepaspectratio]{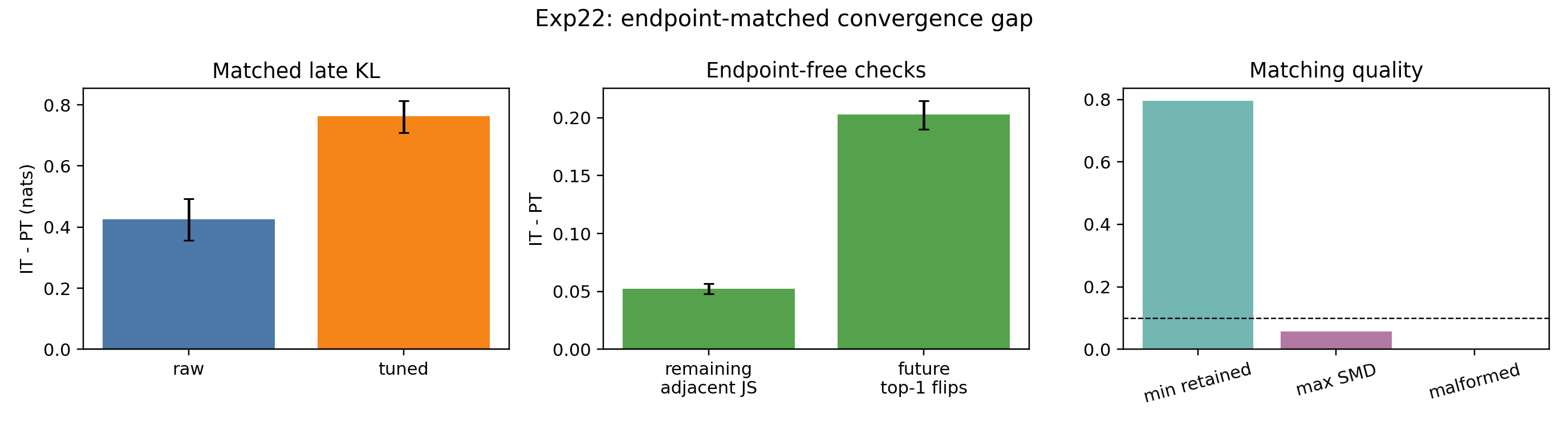}
\caption{Endpoint-matched convergence-gap controls. Matching endpoint confidence and entropy does not remove the IT-minus-PT late gap, and endpoint-free stability checks remain positive.}
\end{figure}

Because instruction-tuned checkpoints are released for serialized chat interfaces, native chat templates are the main prompting condition rather than a cosmetic wrapper. Appendix A removes free-generation length confounding by generating one IT-native continuation per prompt and replaying the same forced token IDs through PT raw, IT native-chat, and IT raw/no-template cells. Under this fixed history, same-prompt late-KL effects are +1.181 nats for native-chat IT versus PT raw and +0.549 for raw/no-template IT versus PT raw; the native-minus-raw template delta is +0.632. Thus raw/no-template serialization attenuates the signature but does not erase it, and the native-template result survives forced continuation replay.

\begin{figure}[H]
\centering
\includegraphics[width=0.94\linewidth,height=0.20\textheight,keepaspectratio]{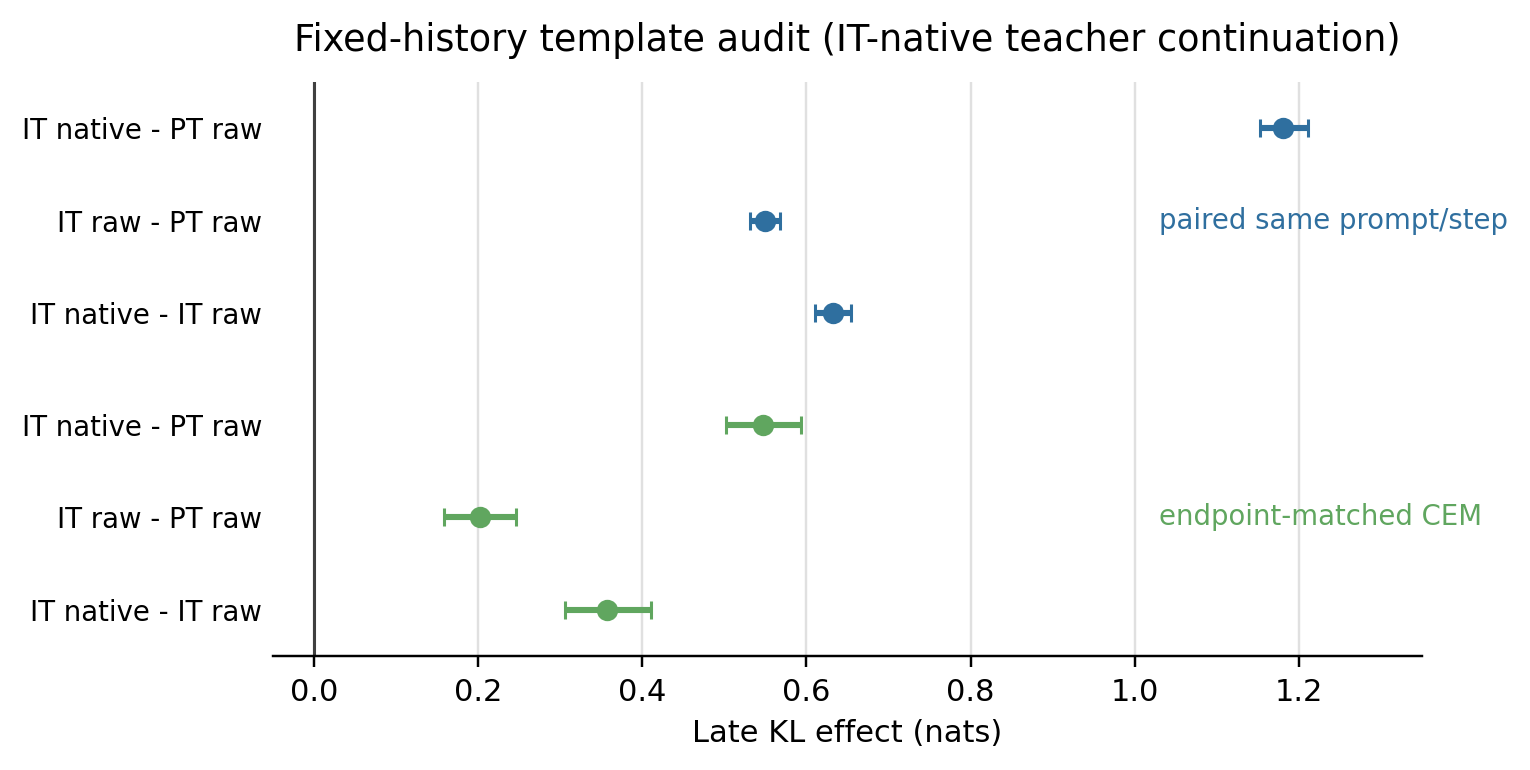}
\caption{Fixed-history template audit. The native-chat effect survives forced continuation replay; raw/no-template replay remains positive but attenuated.}
\end{figure}

\subsection{Late MLP Windows Have the Largest Tested Bidirectional Leverage}\label{late-mlp-windows-have-the-largest-tested-bidirectional-leverage}

Native curves establish the phenomenon; matched-prefix interventions test where the delayed-stabilization signature is easiest to move. On the PT host, replacing late PT MLPs with IT MLPs recreates the delay much more than early or middle grafts. In the dense-family mean, final-20\% KL changes by +0.34 nats for the late graft, compared with -0.03 early and -0.05 middle.

\begin{figure}[H]
\centering
\includegraphics[width=0.94\linewidth,height=0.20\textheight,keepaspectratio]{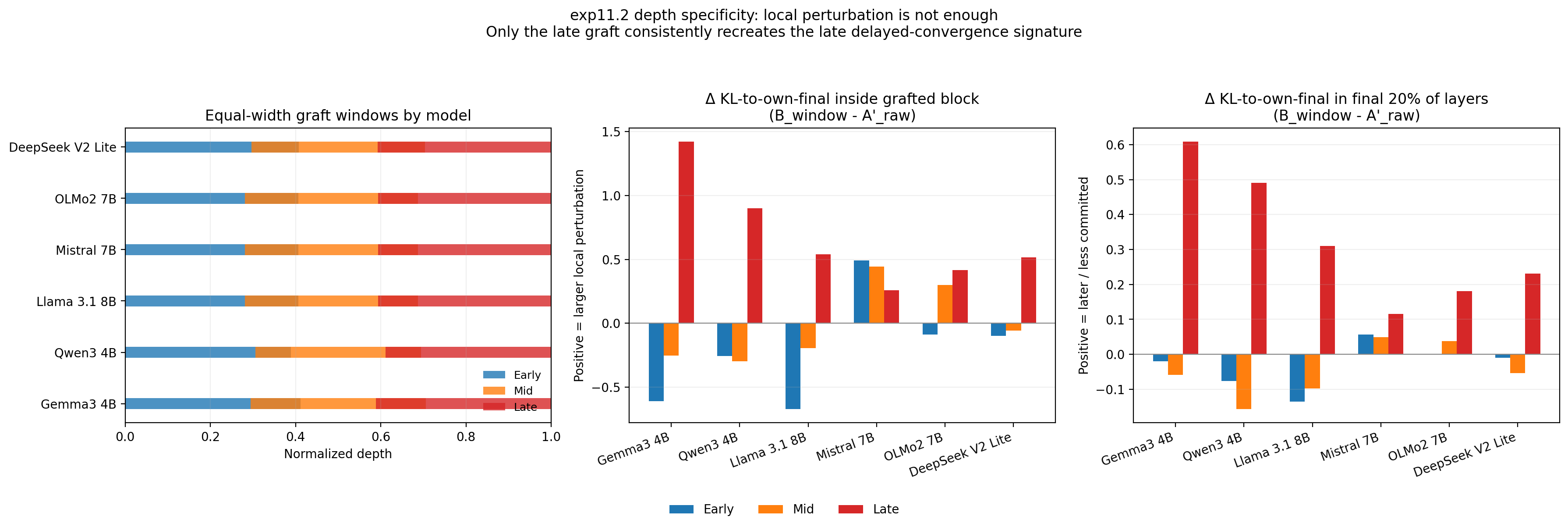}
\caption{Depth-specific matched-prefix IT-MLP grafts into PT hosts. Late grafts most strongly recreate the delayed-stabilization signature.}
\end{figure}

The mirrored IT-host swap points to the same window. Replacing IT late MLPs with PT late MLPs reduces the IT delay by -0.51 nats, compared with -0.10 early and -0.23 middle. The same late window is therefore the largest tested bidirectional MLP handle on the convergence-gap metric.

\begin{figure}[H]
\centering
\includegraphics[width=0.94\linewidth,height=0.20\textheight,keepaspectratio]{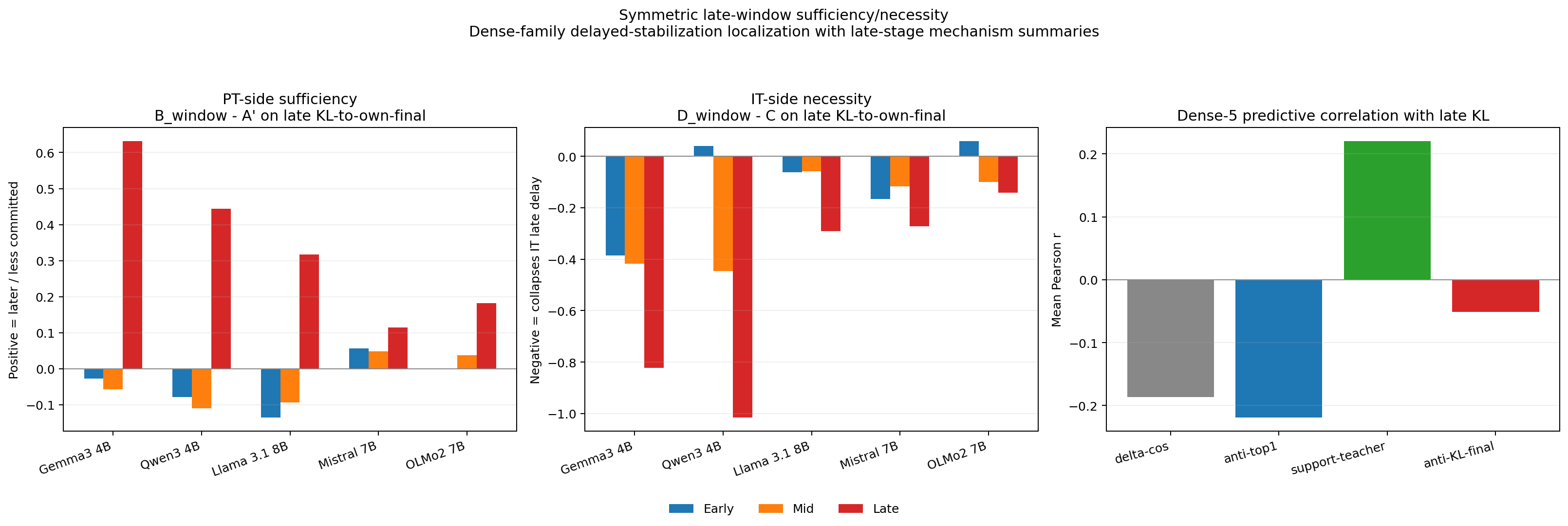}
\caption{Symmetric matched-prefix graft/swap leverage. Late MLPs are the largest tested bidirectional window for delayed stabilization.}
\end{figure}

{\def\LTcaptype{none} 
\begin{longtable}[]{@{}lrrr@{}}
\toprule\noalign{}
Intervention & Early & Middle & Late \\
\midrule\noalign{}
\endhead
\bottomrule\noalign{}
\endlastfoot
IT MLP graft into PT host & \texttt{-0.03} & \texttt{-0.05} & \texttt{+0.34} \\
PT MLP swap into IT host & \texttt{-0.10} & \texttt{-0.23} & \texttt{-0.51} \\
\end{longtable}
}

The sign is not the important part for both rows; the matched direction is. IT late grafts increase the IT-like delay in PT hosts, while PT late swaps reduce the IT delay in IT hosts. Both directions identify late MLPs as the strongest tested window for moving the stabilization trajectory. Figure 5 also makes the family heterogeneity visible: Gemma and Qwen are largest, but all five dense families have positive late-graft and negative late-swap point estimates.

Appendix B stress-tests this interpretation with a late-window width/center audit. The swap side remains stable at the late-full final-20\% window (-0.625 nats, CI excluding zero), while the downstream final-20\% late-full graft is smaller (+0.070, CI crossing zero) and the pre-late swap control also moves. The edited late-window graft effect is clearer (+0.365, CI excluding zero). We therefore read the intervention evidence as late-window leverage on the convergence-gap metric, not as proof of a sharp late-only module or a complete circuit.

\subsection{Matched Random Late Perturbations Do Not Reproduce the Effect}\label{matched-random-late-perturbations-do-not-reproduce-the-effect}

Late layers are close to the unembedding, so a skeptical explanation is that any sufficiently large late perturbation changes convergence. We test that directly with matched random residual-projection controls. The Exp19 control uses 120 prompts per dense family and three late random seeds; the true late graft changes final-20\% KL by +0.327 nats, while the matched random late perturbation gives only +0.003 nats.

\begin{figure}[H]
\centering
\includegraphics[width=0.94\linewidth,height=0.20\textheight,keepaspectratio]{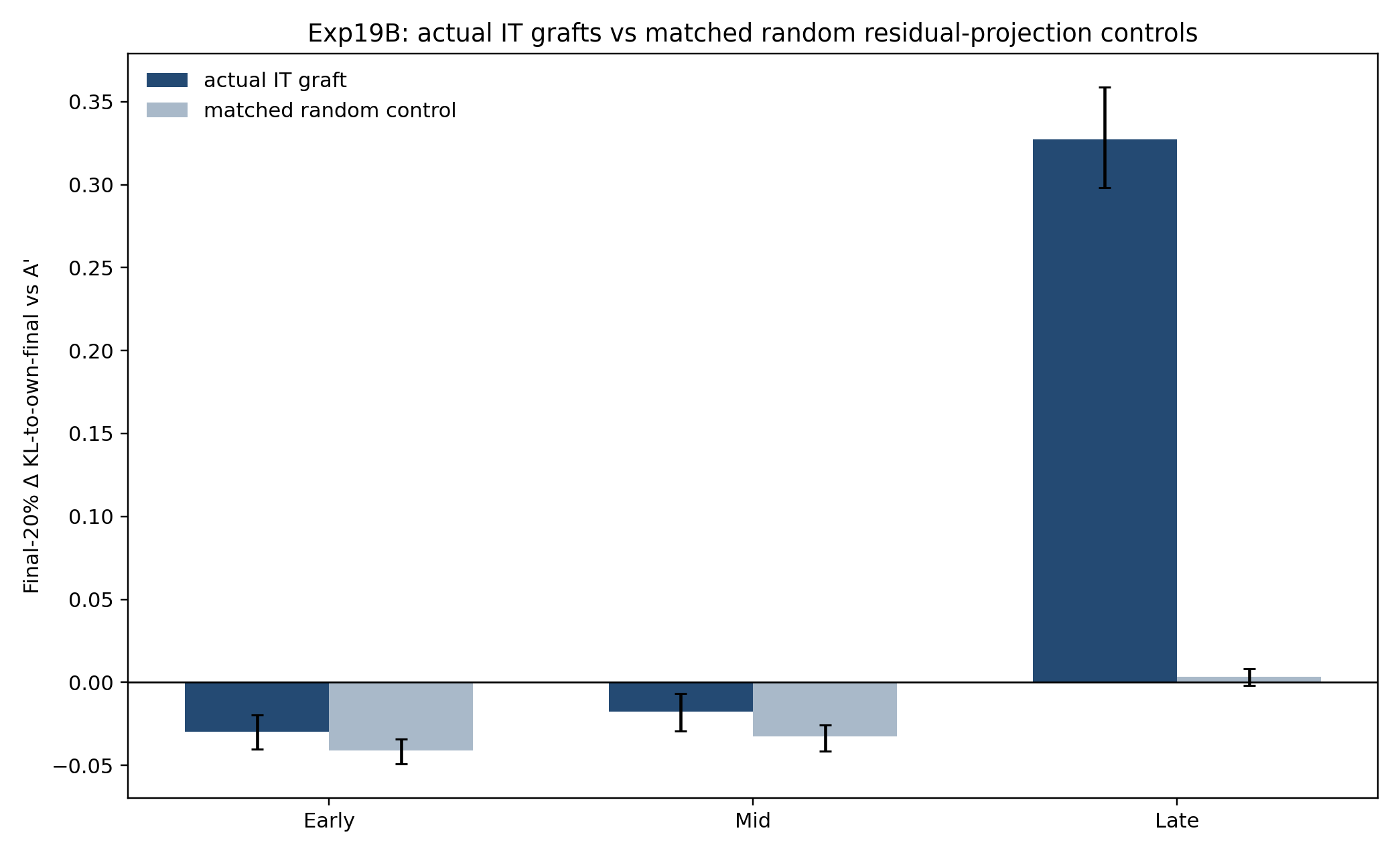}
\caption{True late grafts versus matched random late residual-projection controls. Random perturbations do not reproduce the delayed-stabilization signature.}
\end{figure}

This does not prove that the real update is uniquely efficient or that every late random baseline is exhausted. It rules out the simplest magnitude-only account within this control family: the observed delayed-stabilization effect is not reproduced by this matched random late residual-projection perturbation family.

\subsection{Gemma Behavioral Case Study}\label{gemma-behavioral-case-study}

The convergence gap is not itself a behavior metric. To check whether the same late-window handle can matter under natural decoding, we add an illustrative Gemma case study using the matched-prefix intervention family. Gemma is selected before this paper-facing analysis because it has the largest dense-family bidirectional late-window effects in Appendix B (+0.609 late graft, -0.822 late swap). This is a case study, not a cross-family behavioral claim.

Under free-running 512-token generation on the Exp15 600-prompt support, we compare native Gemma IT (\texttt{C\_it\_chat}) to the same IT host with a late PT MLP swap (\texttt{D\_late\_ptswap}). Blind pairwise judging prefers native IT over the late PT swap on assistant-register prompts in 94.7\% of cases {[}91.6\%, 97.3\%{]} and on safety/format prompts in 93.3\% {[}86.7\%, 98.7\%{]}. The pointwise assistant-register score drops from 4.733 to 3.929, while the early PT swap leaves that score essentially unchanged (4.751). Programmatic content checks do not show a matching collapse: MMLU forced-choice accuracy changes from 0.517 to 0.483, and reasoning exact match changes from 0.850 to 0.925.

\begin{figure}[H]
\centering
\includegraphics[width=0.94\linewidth,height=0.20\textheight,keepaspectratio]{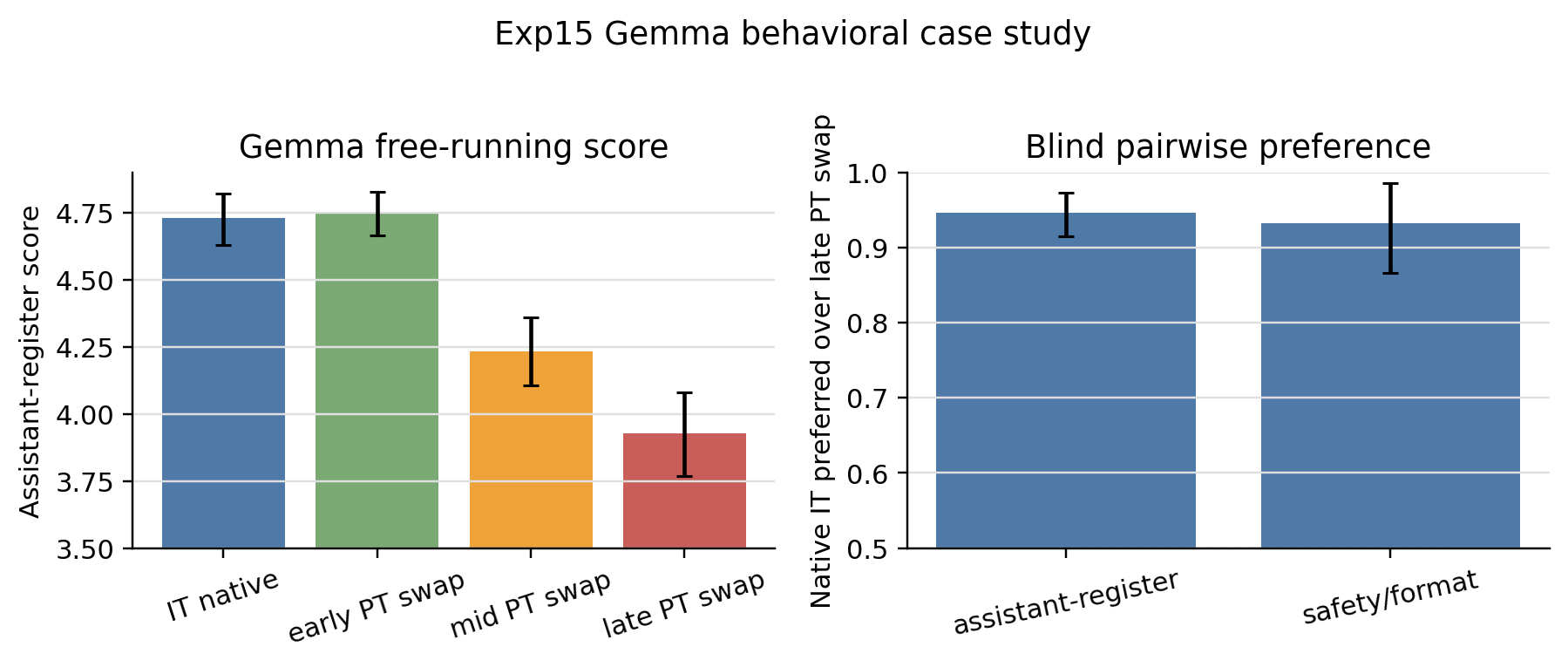}
\caption{Gemma behavioral case study. Native IT is strongly preferred over a late PT MLP swap under free-running generation, while early swaps preserve the assistant-register score.}
\end{figure}

This result is best read as a behavior-facing bridge for the clearest family. It tests whether the late-window intervention that moves the convergence-gap metric can also matter downstream under generation, while the load-bearing claim remains the layerwise stabilization result. It does not establish that the convergence gap predicts broad instruction-following behavior across families or benchmarks.

\section{Discussion}\label{discussion}

\subsection{Prediction Dynamics Rather Than Final Scores}\label{prediction-dynamics-rather-than-final-scores}

The convergence gap reframes post-training as a change in the route to a prediction, not only the endpoint. The key observation is not that late layers sharpen logits; it is that released instruction-following checkpoints remain farther from their own final distribution later in the stack, even after endpoint matching and under endpoint-free same-history checks.

\subsection{What the MLP Interventions Show}\label{what-the-mlp-interventions-show}

The matched-prefix graft/swap results are targeted counterfactual tests on a stabilization metric. They isolate which tested MLP window has the most leverage on the delayed-stabilization signature under matched histories, without requiring a full account of how the host model builds the upstream state. This is the right interpretation of the causal evidence: late MLP windows are the largest tested handle on the effect, but the result is still distribution- and metric-specific.

Just as important is what the intervention does not show. It does not say that late MLPs are a self-contained instruction-following module, that earlier layers are irrelevant, or that the late window is a sharp architectural boundary. A later-stabilizing trajectory can still depend on upstream representations; this paper isolates a prediction-dynamics effect rather than reconstructing the full circuit that produces it.

\subsection{Relation to Prior Work}\label{relation-to-prior-work}

The convergence gap builds on lens-based prediction tracing, but it uses that tracing as a paired post-training diagnostic. Prior layerwise work asks what latent predictions look like or how earlier and later logits differ; here the comparison is between matched PT/IT checkpoint pairs and each model's own endpoint. The diagnostic is useful because it turns ``when does the model settle?'' into a reproducible quantity that can be endpoint-matched, replayed under fixed histories, and targeted by interventions.

{\def\LTcaptype{none} 
\begin{longtable}[]{@{}
  >{\raggedright\arraybackslash}p{(\linewidth - 4\tabcolsep) * \real{0.3333}}
  >{\raggedright\arraybackslash}p{(\linewidth - 4\tabcolsep) * \real{0.3333}}
  >{\raggedright\arraybackslash}p{(\linewidth - 4\tabcolsep) * \real{0.3333}}@{}}
\toprule\noalign{}
\begin{minipage}[b]{\linewidth}\raggedright
Line of work
\end{minipage} & \begin{minipage}[b]{\linewidth}\raggedright
Main target
\end{minipage} & \begin{minipage}[b]{\linewidth}\raggedright
What is added here
\end{minipage} \\
\midrule\noalign{}
\endhead
\bottomrule\noalign{}
\endlastfoot
Tuned/logit lenses & decode latent predictions through depth & paired PT/IT convergence to each model's own final distribution \\
Layer-contrast decoding and stages & exploit or describe differences between earlier and later layers & post-training diagnostic for when predictions stabilize \\
Feed-forward vocabulary updates & interpret MLPs as additive prediction updates & matched-prefix MLP graft/swap leverage on stabilization \\
Instruction-tuning model diffs & study behavioral and representational shifts after instruction tuning & depthwise prediction-stabilization signature across released PT/IT pairs \\
Activation patching and causal tracing & localize components for chosen output metrics & counterfactual tests on the model's stabilization trajectory \\
\end{longtable}
}

The layer location also fits recent instruction-tuning representation studies. Zhao, Ziser, and Cohen divide Llama 2 SFT into shared early layers, transition layers, and refinement layers, with their refinement range covering layers 16-32; our dense-family late windows such as Llama 19-31, Mistral 19-31, OLMo 19-31, Gemma 20-33, and Qwen 22-35 overlap that refinement regime. The overlap is suggestive rather than definitive: Zhao et al.~study task-specific representation similarity, while this paper measures next-token prediction stabilization and interventions on matched histories.

Feature-level model-diffing work is the natural next rung down in granularity. Crosscoders learn shared sparse dictionaries across layers or models, and recent chat-tuning crosscoder work argues that some chat-specific latents are interpretable and causally effective while also warning about sparsity artifacts. The convergence gap is coarser than such feature-level analyses, but it provides a diagnostic target: if a future crosscoder or SAE explains the delayed-stabilization signature, it would connect the layerwise phenomenon here to named latents and downstream behaviors.

This positioning keeps the claim scoped. The convergence gap is not a direct estimate of behavior, and the graft/swap results do not by themselves identify a named sparse feature or full circuit. They show a robust diagnostic signature and the largest tested MLP handle for moving that signature under the evaluated distributions.

\subsection{Limitations}\label{limitations}

The main limitations are probe dependence, intervention scope, template regime, and model-family scope. Tuned-lens curves are estimators, so we pair them with raw-lens and endpoint-free checks. Grafts and swaps are constructed matched-prefix interventions, not natural deployment trajectories.

For instruction-tuned checkpoints, chat templates are part of the released trained interface: they serialize the conversation format and control tokens under which the models are normally optimized and used. Appendix A reports dense-family fixed-history audits: native-template replay preserves the late gap, raw/no-template replay remains positive but attenuated, and a PT-teacher reverse replay has the same sign but weaker template separation with imperfect endpoint balance. We therefore treat no-template prompting as an out-of-distribution serialization that modulates expression of the gap, not as a replacement for the trained IT interface.

Appendix B adds a late-window width/center audit: the late-full swap effect is stable, but the graft-side final-window estimate is small and the pre-late swap control also moves. We therefore claim late MLPs are the strongest tested bidirectional handle, not a sharp layer boundary or a self-contained module. DeepSeek-V2-Lite is included in discovery curves but not pooled into dense-family causal averages because MoE routing changes the intervention interpretation. Finally, the paper does not identify a named sparse feature circuit; it identifies a robust layerwise signature, the largest tested MLP window for moving that signature, and one behavior-facing case study. A natural next step is broader behavioral validation: testing whether late-gap magnitude predicts instruction following, refusal, format adherence, calibration, or prompt sensitivity across families.

\section{Conclusion}\label{conclusion}

Instruction-tuned checkpoints do not merely end at different next-token distributions; they tend to settle toward those distributions later in the forward pass. The convergence gap makes that prediction-dynamics change measurable. Across six PT/IT pairs, the gap survives endpoint matching, endpoint-free same-history controls, and fixed-history template replay. Matched-prefix grafts and swaps identify late MLP windows as the largest tested leverage point on the delayed-stabilization signature, matched random late perturbations do not reproduce it, and a preselected Gemma case study shows behavior-facing consequences for the same late PT swap.

The resulting claim is compact and distinct: \textbf{instruction-tuned checkpoints in this sample stabilize later, and late MLPs are the strongest tested bidirectional handle on that delay}.

\clearpage
\section*{References}

Belrose, N., et al.~(2023). Eliciting Latent Predictions from Transformers with the Tuned Lens. arXiv:2303.08112.

Chuang, Y., et al.~(2024). DoLA: Decoding by Contrasting Layers Improves Factuality. \emph{ICLR 2024}.

Du, H., et al.~(2025). How Post-Training Reshapes LLMs: A Mechanistic View on Knowledge, Truthfulness, Refusal, and Confidence. \emph{COLM 2025}.

Geva, M., Schuster, R., Berant, J., and Levy, O. (2022). Transformer Feed-Forward Layers Are Key-Value Memories. \emph{EMNLP 2022}.

Geva, M., Caciularu, A., Wang, K. R., and Goldberg, Y. (2022). Transformer Feed-Forward Layers Build Predictions by Promoting Concepts in the Vocabulary Space. \emph{EMNLP 2022}.

Heimersheim, S., and Nanda, N. (2024). How to Use and Interpret Activation Patching. arXiv:2404.15255.

Joshi, A., Ahmad, A., and Modi, A. (2025). Calibration Across Layers: Understanding Calibration Evolution in LLMs. \emph{EMNLP 2025}.

Lad, V., Lee, J. H., Gurnee, W., and Tegmark, M. (2025). The Remarkable Robustness of LLMs: Stages of Inference? \emph{NeurIPS 2025}.

Lindsey, J., Templeton, A., Marcus, J., Conerly, T., Batson, J., and Olah, C. (2024). Sparse Crosscoders for Cross-Layer Features and Model Diffing. \emph{Transformer Circuits Thread}.

Minder, J., Dumas, C., Juang, C., Chugtai, B., and Nanda, N. (2025). Robustly Identifying Concepts Introduced During Chat Fine-Tuning Using Crosscoders. arXiv:2504.02922.

Prakash, N., Shaham, T. R., Haklay, T., Belinkov, Y., and Bau, D. (2024). Fine-Tuning Enhances Existing Mechanisms: A Case Study on Entity Tracking. \emph{ICLR 2024}.

Wu, X., Yao, W., Chen, J., Pan, X., Wang, X., Liu, N., and Yu, D. (2024). From Language Modeling to Instruction Following: Understanding the Behavior Shift in LLMs after Instruction Tuning. \emph{NAACL 2024}.

Zhao, Z., Ziser, Y., and Cohen, S. B. (2024). Layer by Layer: Uncovering Where Multi-Task Learning Happens in Instruction-Tuned Large Language Models. \emph{EMNLP 2024}.

\clearpage
\section*{Appendix Roadmap}

{\def\LTcaptype{none} 
\begin{longtable}[]{@{}
  >{\raggedright\arraybackslash}p{(\linewidth - 4\tabcolsep) * \real{0.3333}}
  >{\raggedright\arraybackslash}p{(\linewidth - 4\tabcolsep) * \real{0.3333}}
  >{\raggedright\arraybackslash}p{(\linewidth - 4\tabcolsep) * \real{0.3333}}@{}}
\toprule\noalign{}
\begin{minipage}[b]{\linewidth}\raggedright
Appendix
\end{minipage} & \begin{minipage}[b]{\linewidth}\raggedright
Supports
\end{minipage} & \begin{minipage}[b]{\linewidth}\raggedright
Does not prove
\end{minipage} \\
\midrule\noalign{}
\endhead
\bottomrule\noalign{}
\endlastfoot
A. Convergence-gap robustness & raw/tuned discovery, endpoint matching, and endpoint-free same-history controls & deployment-level behavior \\
B. Matched-prefix MLP leverage & late graft/swap effects, random controls, and width/center robustness & full circuit recovery or a sharp late-only boundary \\
C. Gemma behavioral case study & free-running effect of the late PT swap on one preselected high-signal family & cross-family behavioral generalization \\
D. Reproducibility map & CPU claim checks and artifact hashes & hardware-free full raw reruns \\
\end{longtable}
}

\appendix

\section{Convergence-Gap Robustness}\label{convergence-gap-robustness}

The convergence-gap claim is carried by a compact ladder: native discovery curves, endpoint-matched late KL, endpoint-free adjacent JS, and endpoint-free future top-1 flips. Commitment thresholds are reported as summaries of the same trajectory and are not used as a single decisive definition.

The endpoint-matched control matches token steps by final confidence, entropy, and top1/top2 margin. Minimum matched retention is 0.796; maximum post-match standardized mean difference is 0.057. The remaining IT-minus-PT late gap is therefore not a simple artifact of more confident or less confident endpoints.

Template-regime fixed-history audit: instruction-tuned checkpoints are normally trained and released for a serialized chat interface, so native chat-template prompting is the main condition rather than an optional wrapper. To remove rollout-length confounding, we generated one IT-native continuation per prompt and replayed the same forced token IDs through \texttt{pt\_raw}, \texttt{it\_native}, and \texttt{it\_raw} cells for the four dense families with completed IT-native replay records. Same-prompt/step raw-lens late-KL effects are +1.181 nats {[}+1.153, +1.211{]} for \texttt{it\_native\ -\ pt\_raw}, +0.549 {[}+0.531, +0.568{]} for \texttt{it\_raw\ -\ pt\_raw}, and +0.632 {[}+0.611, +0.654{]} for \texttt{it\_native\ -\ it\_raw}. Endpoint-matched CEM effects are +0.548 {[}+0.502, +0.594{]}, +0.202 {[}+0.158, +0.247{]}, and +0.357 {[}+0.306, +0.411{]}, respectively, with 0.999 minimum retention, 0.061 maximum SMD, zero malformed records, and zero missing aligned token steps. The native-template gap therefore survives fixed-continuation replay; raw/no-template serialization remains positive but attenuated, so it is robustness and limitation evidence rather than a replacement for the trained IT interface.

As a reverse teacher-source stress test, we also generated PT-raw continuations and replayed those histories through the same three cells. The paired raw-lens late-KL effects have the same sign but are weaker: +0.610 {[}+0.135, +1.079{]} for \texttt{it\_native\ -\ pt\_raw}, +0.429 {[}+0.155, +0.639{]} for \texttt{it\_raw\ -\ pt\_raw}, and +0.181 {[}-0.105, +0.467{]} for \texttt{it\_native\ -\ it\_raw}. Collection quality is clean, with zero malformed records and zero missing aligned steps, but endpoint balance is not uniformly within the pre-specified SMD threshold (0.991 minimum retention, 0.155 maximum SMD). We therefore use this reverse replay only as same-sign support under PT histories, not as stronger evidence for the template-delta claim.

The fixed-history replay table is generated from the checked synthesis artifacts:

{\def\LTcaptype{none} 
\begin{longtable}[]{@{}
  >{\raggedright\arraybackslash}p{(\linewidth - 10\tabcolsep) * \real{0.1429}}
  >{\raggedright\arraybackslash}p{(\linewidth - 10\tabcolsep) * \real{0.1429}}
  >{\raggedleft\arraybackslash}p{(\linewidth - 10\tabcolsep) * \real{0.1905}}
  >{\raggedleft\arraybackslash}p{(\linewidth - 10\tabcolsep) * \real{0.1905}}
  >{\raggedleft\arraybackslash}p{(\linewidth - 10\tabcolsep) * \real{0.1905}}
  >{\raggedright\arraybackslash}p{(\linewidth - 10\tabcolsep) * \real{0.1429}}@{}}
\toprule\noalign{}
\begin{minipage}[b]{\linewidth}\raggedright
Teacher history
\end{minipage} & \begin{minipage}[b]{\linewidth}\raggedright
Estimator
\end{minipage} & \begin{minipage}[b]{\linewidth}\raggedleft
IT native - PT raw
\end{minipage} & \begin{minipage}[b]{\linewidth}\raggedleft
IT raw - PT raw
\end{minipage} & \begin{minipage}[b]{\linewidth}\raggedleft
IT native - IT raw
\end{minipage} & \begin{minipage}[b]{\linewidth}\raggedright
Use/quality
\end{minipage} \\
\midrule\noalign{}
\endhead
\bottomrule\noalign{}
\endlastfoot
IT-native continuation & paired same prompt/step & \texttt{+1.181} \texttt{{[}+1.153,\ +1.211{]}} & \texttt{+0.549} \texttt{{[}+0.531,\ +0.568{]}} & \texttt{+0.632} \texttt{{[}+0.611,\ +0.654{]}} & primary fixed-history replay \\
IT-native continuation & endpoint-matched CEM & \texttt{+0.548} \texttt{{[}+0.502,\ +0.594{]}} & \texttt{+0.202} \texttt{{[}+0.158,\ +0.247{]}} & \texttt{+0.357} \texttt{{[}+0.306,\ +0.411{]}} & retention \texttt{0.999}; max SMD \texttt{0.061} \\
PT-raw continuation & paired same prompt/step & \texttt{+0.610} \texttt{{[}+0.135,\ +1.079{]}} & \texttt{+0.429} \texttt{{[}+0.155,\ +0.639{]}} & \texttt{+0.181} \texttt{{[}-0.105,\ +0.467{]}} & same-sign reverse stress test \\
PT-raw continuation & endpoint-matched CEM & \texttt{+0.025} \texttt{{[}-0.183,\ +0.233{]}} & \texttt{+0.012} \texttt{{[}-0.040,\ +0.064{]}} & \texttt{-0.106} \texttt{{[}-0.274,\ +0.111{]}} & balance caveat: retention \texttt{0.991}; max SMD \texttt{0.155} \\
\end{longtable}
}

Discovery-curve token-step counts:

{\def\LTcaptype{none} 
\begin{longtable}[]{@{}lrrr@{}}
\toprule\noalign{}
Family & PT token steps & IT token steps & Layers \\
\midrule\noalign{}
\endhead
\bottomrule\noalign{}
\endlastfoot
Gemma 3 4B & \texttt{1,273,606} & \texttt{810,347} & \texttt{34} \\
Llama 3.1 8B & \texttt{517,579} & \texttt{499,240} & \texttt{32} \\
Qwen 3 4B & \texttt{482,319} & \texttt{636,979} & \texttt{36} \\
Mistral 7B v0.3 & \texttt{1,380,081} & \texttt{501,093} & \texttt{32} \\
OLMo 2 7B & \texttt{370,171} & \texttt{166,698} & \texttt{32} \\
DeepSeek-V2-Lite & \texttt{184,658} & \texttt{163,877} & \texttt{27} \\
\end{longtable}
}

\section{Matched-Prefix MLP Leverage}\label{matched-prefix-mlp-leverage}

Matched-prefix graft/swap experiments use overlapping early, middle, and late MLP windows. The dense-family causal pool excludes DeepSeek because MoE routing changes the interpretation of swapping dense MLP windows.

The random-control follow-up uses matched late residual-projection perturbations. The true late graft gives +0.327 final-20\% KL change, while the matched random perturbation gives +0.003. This control is the main reason we interpret the late result as a specific matched-prefix update rather than generic late fragility.

Per-family headline late-window effects show that the dense mean is not driven by a single model:

{\def\LTcaptype{none} 
\begin{longtable}[]{@{}
  >{\raggedright\arraybackslash}p{(\linewidth - 6\tabcolsep) * \real{0.2143}}
  >{\raggedleft\arraybackslash}p{(\linewidth - 6\tabcolsep) * \real{0.2857}}
  >{\raggedleft\arraybackslash}p{(\linewidth - 6\tabcolsep) * \real{0.2857}}
  >{\raggedright\arraybackslash}p{(\linewidth - 6\tabcolsep) * \real{0.2143}}@{}}
\toprule\noalign{}
\begin{minipage}[b]{\linewidth}\raggedright
Family
\end{minipage} & \begin{minipage}[b]{\linewidth}\raggedleft
Late IT graft into PT host
\end{minipage} & \begin{minipage}[b]{\linewidth}\raggedleft
Late PT swap into IT host
\end{minipage} & \begin{minipage}[b]{\linewidth}\raggedright
Late window
\end{minipage} \\
\midrule\noalign{}
\endhead
\bottomrule\noalign{}
\endlastfoot
Gemma 3 4B & \texttt{+0.609} & \texttt{-0.822} & \texttt{20-33} \\
Qwen 3 4B & \texttt{+0.491} & \texttt{-1.015} & \texttt{22-35} \\
Llama 3.1 8B & \texttt{+0.310} & \texttt{-0.291} & \texttt{19-31} \\
Mistral 7B v0.3 & \texttt{+0.115} & \texttt{-0.273} & \texttt{19-31} \\
OLMo 2 7B & \texttt{+0.181} & \texttt{-0.142} & \texttt{19-31} \\
Dense mean & \texttt{+0.341} & \texttt{-0.509} & - \\
\end{longtable}
}

The effect is heterogeneous in magnitude but sign-consistent for the late-window headline across the dense families. The current paper therefore treats the dense mean as a center-of-mass summary and lists per-family values for audit.

A late-window width/center audit reruns the matched-prefix experiment over six overlapping windows on the same five dense families (600 prompts, 128 forced steps). The table reports dense-family model-mean \texttt{KL(layer\ \textbar{}\textbar{}\ own\ final)} deltas; final-20\% values measure downstream stabilization, while graft-window values measure the directly edited layers.

{\def\LTcaptype{none} 
\begin{longtable}[]{@{}
  >{\raggedright\arraybackslash}p{(\linewidth - 8\tabcolsep) * \real{0.1579}}
  >{\raggedleft\arraybackslash}p{(\linewidth - 8\tabcolsep) * \real{0.2105}}
  >{\raggedleft\arraybackslash}p{(\linewidth - 8\tabcolsep) * \real{0.2105}}
  >{\raggedleft\arraybackslash}p{(\linewidth - 8\tabcolsep) * \real{0.2105}}
  >{\raggedleft\arraybackslash}p{(\linewidth - 8\tabcolsep) * \real{0.2105}}@{}}
\toprule\noalign{}
\begin{minipage}[b]{\linewidth}\raggedright
Window
\end{minipage} & \begin{minipage}[b]{\linewidth}\raggedleft
Final-20 IT graft into PT
\end{minipage} & \begin{minipage}[b]{\linewidth}\raggedleft
Final-20 PT swap into IT
\end{minipage} & \begin{minipage}[b]{\linewidth}\raggedleft
Edited-window IT graft into PT
\end{minipage} & \begin{minipage}[b]{\linewidth}\raggedleft
Edited-window PT swap into IT
\end{minipage} \\
\midrule\noalign{}
\endhead
\bottomrule\noalign{}
\endlastfoot
Pre-late half & \texttt{-0.001} & \texttt{-0.384} & \texttt{+0.069} & \texttt{-2.674} \\
Late full & \texttt{+0.070} & \texttt{-0.625} & \texttt{+0.365} & \texttt{-1.605} \\
Late front half & \texttt{+0.008} & \texttt{-0.352} & \texttt{+0.142} & \texttt{-1.321} \\
Late center half & \texttt{+0.022} & \texttt{-0.352} & \texttt{+0.094} & \texttt{-0.701} \\
Late terminal half & \texttt{+0.050} & \texttt{-0.443} & \texttt{+0.040} & \texttt{-0.403} \\
Terminal quarter & \texttt{+0.033} & \texttt{-0.347} & \texttt{-0.012} & \texttt{-0.134} \\
\end{longtable}
}

This audit supports the late-window leverage claim but also narrows it. The late-full final-20\% swap remains negative with a bootstrap CI excluding zero (-0.625 nats, {[}-1.076, -0.201{]}), and late subwindows preserve the swap-side sign. The late-full graft is positive but weaker at final-20\% (+0.070, {[}-0.048, +0.189{]}); its within-window effect is clearer (+0.365, {[}+0.108, +0.629{]}), with late-front and late-center graft-window estimates also above zero. Because the pre-late swap control is also negative, the audit should not be read as proving a uniquely late-only boundary. It supports the more conservative claim made in the main text: among tested matched-prefix MLP windows, the late window is the strongest bidirectional handle on delayed stabilization.

\section{Gemma Behavioral Case Study}\label{gemma-behavioral-case-study-1}

The Gemma behavioral case study reuses the same matched-prefix intervention family under free-running generation. It is included to show behavior-facing plausibility for the clearest convergence-intervention family, not to make a population-level behavioral claim. Full dense-family Exp15 behavior is noisier on the PT-graft sufficiency side; the cleanest behavior-facing result is the IT-side late PT swap.

{\def\LTcaptype{none} 
\begin{longtable}[]{@{}
  >{\raggedright\arraybackslash}p{(\linewidth - 8\tabcolsep) * \real{0.1579}}
  >{\raggedleft\arraybackslash}p{(\linewidth - 8\tabcolsep) * \real{0.2105}}
  >{\raggedleft\arraybackslash}p{(\linewidth - 8\tabcolsep) * \real{0.2105}}
  >{\raggedleft\arraybackslash}p{(\linewidth - 8\tabcolsep) * \real{0.2105}}
  >{\raggedleft\arraybackslash}p{(\linewidth - 8\tabcolsep) * \real{0.2105}}@{}}
\toprule\noalign{}
\begin{minipage}[b]{\linewidth}\raggedright
Readout
\end{minipage} & \begin{minipage}[b]{\linewidth}\raggedleft
IT native
\end{minipage} & \begin{minipage}[b]{\linewidth}\raggedleft
Late PT swap
\end{minipage} & \begin{minipage}[b]{\linewidth}\raggedleft
Effect/preference
\end{minipage} & \begin{minipage}[b]{\linewidth}\raggedleft
N
\end{minipage} \\
\midrule\noalign{}
\endhead
\bottomrule\noalign{}
\endlastfoot
Assistant-register score & \texttt{4.733} \texttt{{[}4.631,\ 4.822{]}} & \texttt{3.929} \texttt{{[}3.769,\ 4.080{]}} & \texttt{+0.804} \texttt{{[}+0.649,\ +0.960{]}} & \texttt{225} \\
Pairwise assistant-register & \texttt{0.947} \texttt{{[}0.916,\ 0.973{]}} & \texttt{0.053} \texttt{{[}0.027,\ 0.084{]}} & native preferred & \texttt{225} \\
Pairwise safety/format & \texttt{0.933} \texttt{{[}0.867,\ 0.987{]}} & \texttt{0.067} \texttt{{[}0.013,\ 0.133{]}} & native preferred & \texttt{75} \\
MMLU forced-choice accuracy & \texttt{0.517} & \texttt{0.483} & \texttt{-0.033} & \texttt{60} \\
Reasoning exact match & \texttt{0.850} & \texttt{0.925} & \texttt{+0.075} & \texttt{40} \\
Format compliance & \texttt{0.574} & \texttt{0.412} & \texttt{-0.162} & \texttt{68} \\
\end{longtable}
}

\section{Reproducibility and Artifact Map}\label{reproducibility-and-artifact-map}

We provide a CPU-only reviewer artifact bundle that regenerates the quoted paper numbers from committed JSON/CSV summaries. Full raw intervention reruns require multi-GPU hardware and are optional for audit.

{\def\LTcaptype{none} 
\begin{longtable}[]{@{}
  >{\raggedright\arraybackslash}p{(\linewidth - 2\tabcolsep) * \real{0.5000}}
  >{\raggedright\arraybackslash}p{(\linewidth - 2\tabcolsep) * \real{0.5000}}@{}}
\toprule\noalign{}
\begin{minipage}[b]{\linewidth}\raggedright
Claim group
\end{minipage} & \begin{minipage}[b]{\linewidth}\raggedright
Checker/artifacts
\end{minipage} \\
\midrule\noalign{}
\endhead
\bottomrule\noalign{}
\endlastfoot
Convergence gap & CPU claim checker; reporting-table generator; Exp9/Exp22 summary JSON/CSV; generated reporting tables \\
Matched-prefix MLP leverage & Exp11 depth-ablation metrics; Exp14 symmetric graft/swap summary; Exp19B random-control summary; Exp55 late-window width/center synthesis \\
Gemma behavioral case study & Exp15 Gemma behavior synthesis; generated reporting table \\
\end{longtable}
}

\end{document}